\documentclass{article}

    \usepackage[preprint]{neurips_2020}

\usepackage[utf8]{inputenc} 
\usepackage[T1]{fontenc}    
\usepackage{hyperref}       
\usepackage{url}            
\usepackage{booktabs}       
\usepackage{amsfonts}       
\usepackage{nicefrac}       
\usepackage{microtype}      
\usepackage{graphicx}
\usepackage{float}
\usepackage{amsfonts,amssymb,amsmath}
\usepackage[]{algorithm2e}
\usepackage{color,soul}

\title{
Grassmann Iterative Linear Discriminant Analysis
with Proxy Matrix Optimization}

\author{%
  Navya Nagananda \\
  Department of Computer Engineering\\
  Rochester Institute of Technology\\
  Rochester, NY 14624 \\
  \texttt{nn3264@rit.edu} \\
     \And
 Breton Minnehan \\
  Air Force Research Laboratory \\
  WPAFB, OH, 45433 USA \\
  \texttt{breton.minnehan.1@us.af.mil} \\
   \And
  Andreas Savakis \\
  Department of Computer Engineering \\
  Rochester Institute of Technology \\
   Rochester, NY 14624 \\
  \texttt{Andreas.Savakis@rit.edu} \\
}

\begin{document}

\maketitle

\begin{abstract}
  Linear Discriminant Analysis (LDA) is 
  commonly used
  for dimensionality reduction in pattern recognition and statistics. It is a supervised method that aims to find the most discriminant space 
  of reduced dimension
  that can be further used 
  for classification. In this work, we present a Grassmann Iterative LDA  method (GILDA) that is based on Proxy Matrix Optimization (PMO).
  PMO makes use of automatic differentiation and stochastic gradient descent (SGD) on the Grassmann manifold to arrive at the optimal projection matrix. Our results show that GILDA outperforms the prevailing manifold optimization method.
\end{abstract}

\section{Introduction}
Linear dimensionality reduction is a popular tool used in statistics, machine learning, and signal processing. Linear Discriminant Analysis (LDA) finds the linear projections of high-dimensional data into a low-dimensional space \cite{bishop2006pattern}
and determines the optimal linear decision boundaries in the resulting latent space \cite{cunningham2015linear}. Fishers LDA \cite{fisher1936use, lda_bio} learns the low dimensional space by maximizing the inter-class variability while minimizing the intra-class variability.

LDA is usually solved using the generalized eigenvalue solution, however, this is sub-optimal \cite{cunningham2015linear}. 
A better approach would be to cast LDA as an optimization problem over a matrix manifold \cite{absil2009optimization}. In this paper, we propose the Grassmann Iterative LDA (GILDA) method based on the Proxy Matrix Optimization (PMO) approach \cite{Minnehan_2019_CVPR}. 
PMO makes use of Grassmann manifold (GM) optimization combined with a deep learning framework. Previous works have considered the GM in a neural network for various optimization schemes \cite{gm_cluster, gm_sgd, zhang2018grassmannian}.  We propose GILDA in a PMO framework and demonstrate improved results over the two-step optimization method in \cite{cunningham2015linear}. The main contributions of this work are:
\begin{itemize}
    \item The introduction of a proxy matrix optimization scheme in GILDA to find the optimal projection matrix for LDA.
    \item Our PMO-based GILDA method outperforms the popular two-step optimization procedure in \cite{cunningham2015linear}.
    \item The GILDA framework is suitable for implementation in a neural network with end-to-end training, as it uses automatic differentiation 
    \cite{paszke2017automatic} 
    and SGD 
    \cite{sgd} 
    optimization.
\end{itemize}

\section{Background}
Linear Discriminant Analysis (LDA) projects labeled data
in a lower dimensional space, in a way that maximizes the separation between classes. Let $\textbf{X} = [x_1, x_2, \dots , x_N] \in \mathbb{R}^{m \times N}$ be the data matrix which is $m$ dimensional and has $N$ data points. To find the LDA projection, the between class scatter matrix ($\mathbf{\Sigma_B}$) and the within class scatter matrix ($\mathbf{\Sigma_W}$) are calculated as:
\begin{equation}
    \mathbf{\Sigma_W} = \sum_{i=1}^{n} (x_i - \mu_{c_i})(x_i - \mu_{c_i})^T \quad\text{ and }\quad \mathbf{\Sigma_B} = \sum_{i=1}^{n} (\mu_{c_i} - \mu)(\mu_{c_i} - \mu)^T,
    \label{eq:lda_scatter}
\end{equation}
where $\mu$ is the mean of the entire dataset and $\mu_{c_i}$ is the class mean associated with $x_i$. The LDA projection matrix 
$\textbf{R} \in \mathbb{O}^{m \times p}$ ($p<m$ is the dimension of the lower dimensional space) aims to maximize the between-class variability while minimizing the within-class variability, which leads to minimizing the following objective,
\begin{equation}
    f = -\frac{trace(\textbf{R}^T \mathbf{\Sigma_B} \textbf{R})}{trace(\textbf{R}^T \mathbf{\Sigma_W} \textbf{R})}.
    \label{eq:lda_obj}
\end{equation}
The projection matrix $\textbf{R}$ is orthogonal
under this objective function.
The eigenvalue solution considers the top $p$ eigenvectors of the objective ($\mathbf{\Sigma_W}^{-1} \mathbf{\Sigma_B}$).

\section{Manifold Optimization}
Manifolds are used in machine learning for dimensionality reduction \cite{edelman1998geometry} and other applications. 
The Grassmann manifold $\mathcal{G}^{m \times p}$ \cite{hamm_gm, turaga_gm} is of particular interest, because every point on $\mathcal{G}^{m \times p}$ is a linear subspace specified by an
orthogonal basis represented by $m \times p$ dimensional matrices \cite{edelman1998geometry}.
\begin{equation*}
    \mathcal{G}^{m \times p} \overset{\Delta}{=} \{Span(\textbf{X}): \textbf{X} \in \mathbb{R}^{m \times p}, \textbf{X}^T \textbf{X} = \textbf{I}_p\}
\end{equation*}
Two points on $\mathcal{G}^{m \times p}$
are equivalent if their columns span the same $p$-dimensional subspace \cite{edelman1998geometry}. 
The tangent space of the manifold $M$ at a point $\textbf{Y} \in M$ is the linear approximation of the manifold at a particular point and contains all the tangent vectors to $M$ at $\textbf{Y}$. Mathematically, the tangent space, $T_YM$ is the set of all points $\textbf{X}$ that satisfy $\textbf{Y}^T\textbf{X} + \textbf{X}^T \textbf{Y} = \textbf{0}_p$, where $\textbf{0}_p$ is the matrix containing all zero entries. Another important geometric concept is the geodesic, which is
the shortest length connecting two points on the manifold.

When optimizing over the Grassmann manifold, the direction of each update step is found in the tangent space of the current location on the manifold. For a given objective $F$, the gradient with respect to a point on the manifold $\textbf{Y} \in M$ is given by:
\begin{equation}
    \nabla F = \frac{\partial F}{\partial \mathbf{Y}} - \mathbf{Y} (\frac{\partial F}{\partial \mathbf{Y}})^{T}\mathbf{\mathbf{Y}}^T.
    \label{eq:man_grad}
\end{equation}
The gradient of the loss however is not restricted to the manifold of the tangent space. Thus, it needs to be projected back onto the tangent space. The equation for the projection of a point $\textbf{Z}$ in ambient Euclidean space to the tangent space of the manifold defined at point $\textbf{Y}$ is given by:
\begin{equation}
    \pi_{T,{\mathbf{Y}}}({\mathbf{Z}}) = \mathbf{Y}\frac{1}{2}(\mathbf{Y}^{T}\mathbf{Z}-\mathbf{Z}^{T}\mathbf{Y})+(\mathbf{I}_{p}-\mathbf{Y}\mathbf{Y}^{T})\mathbf{Z}.
    \label{eq:tanproj}
\end{equation}
This operation eliminates the components normal to the tangent space while preserving the components on the tangent space of the manifold at point $\textbf{Y}$.
The retraction $r_p$ is a mapping from a point in ambient space to the closest point on the manifold:
$r_p: T_p M \rightarrow M$. In this work we use the retraction operation defined by \cite{cunningham2015linear} which makes use of the SVD of $\textbf{Z}$, $\textbf{Z} = \textbf{U} \mathbf{\Sigma} \textbf{V}^T$ as:
\begin{equation}
   r_{\mathbf{p}}(\mathbf{Z}) = \mathbf{U}\mathbf{V}^{T}.
   \label{eq:retract}
\end{equation}

The two-step approach for retraction from ambient space on the manifold is used by \cite{cunningham2015linear} to find the optimal projection matrix. 
A point $\textbf{R}_i$, at iteration $i$, is updated based on the optimization gradients that produce $\textbf{Z}_{i+1}$ in ambient Euclidean space at the next iteration. 
The point $\textbf{Z}_{i+1}$ is projected to the tangent space using Eq. \ref{eq:tanproj}.
Following the projection to tangent space, the point is then retracted to the manifold using Eq. \ref{eq:retract}, resulting in point $\textbf{R}_{i+1}$ on the manifold. 


\subsection{Proxy Matrix Optimization}
The PMO approach in GILDA performs the optimization steps in  ambient space, such that each update is in the loss minimizing direction. 
PMO embeds the manifold retraction inside the optimization function, instead of performing retraction after optimization. 
Unlike the two-step process, the PMO search is not restricted to the local region of the manifold, and therefore, the extent of the optimization step is not limited. 
Thus,  PMO 
allows greater steps in ambient space and achieves faster convergence.

\begin{figure}[H]
\centering
\includegraphics[width = 5cm]{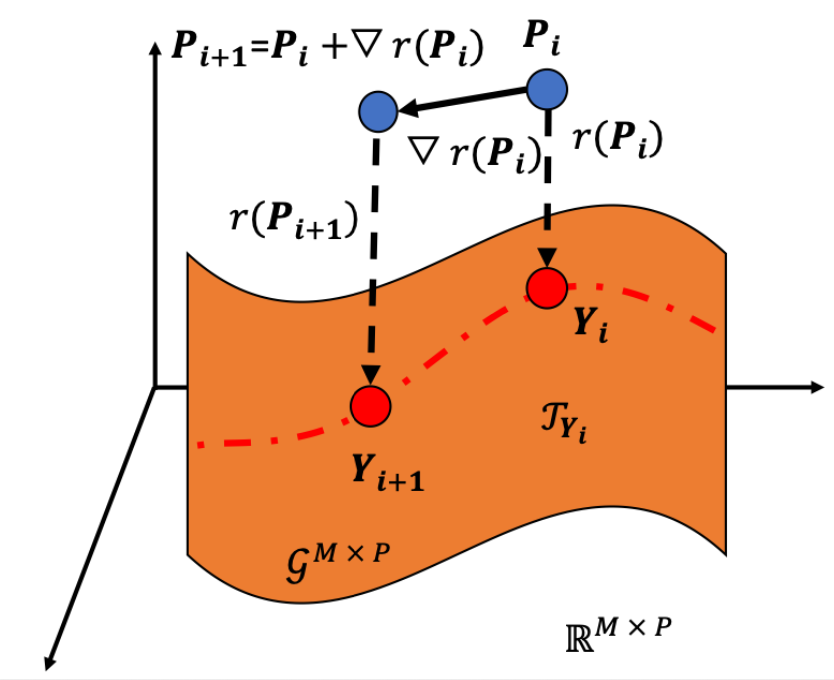}
\caption{Illustration of the proxy matrix optimization procedure.}
\label{fig:proxy}
\end{figure}

The PMO method does not aim to directly optimize a matrix on the manifold, but
instead it uses an auxiliary, or Proxy Matrix, that exists in the ambient Euclidean space and is retracted to the closest location on the manifold using Eq. \ref{eq:retract}. The PMO process is illustrated in Fig. \ref{fig:proxy} and the corresponding steps are outlined in Algorithm \ref{alg:proxy}. The first step in the PMO process is to retract the proxy matrix, $\mathbf{P_i}$ to $\mathbf{Y_i}$, its closest location on the manifold. Once the proxy matrix is retracted to the manifold, the loss is calculated based on the loss function at $\mathbf{Y_i}$. This loss is then backpropagated through the singular value decomposition of proxy matrix using a method developed by \cite{ionescu2015training} to a new point $\mathbf{P_{i+1}}$. This point is then retracted back onto the manifold using Eq. \ref{eq:retract} to point $\mathbf{Y_{i+1}}$.

\begin{algorithm}
 \KwData{$\mathbf{X} \in \mathbb{R}^{m \times N}$}
 \KwResult{Locally Optimal $\mathbf{P}$ such that $r(\mathbf{P_i})$ minimizes the loss 
 $f_{\mathbf{X}}$} 
 initialize $\mathbf{P} \in \mathcal{R}^{m \times p}$ and  $\mathbf{P} \not\in \mathcal{G}^{m \times p}$\;
 \small{
 \For{ $i > iter$}{
  $\mathbf{U} \mathbf{S} \mathbf{V}^T = \mathbf{P_i}$
   \tcc*{Retract $\mathbf{P_i}$ to $\mathcal{G}^{m \times p}$} 
   $r(\mathbf{P_i}) = \mathbf{U} \mathbf{V}^T $\;
   $\nabla r(\mathbf{P_i}) = \frac{\partial}{\partial \mathbf{P_i}} f_X(r(\mathbf{P_i}))$
   \tcc*{Calculate gradients for $\mathbf{P_i}$ using Eq.\ref{eq:man_grad}}
   $\mathbf{P_{i+1}}=\mathbf{P_{i}}-\mathbf{\beta\nabla r (P_{i})}$
   \tcc*{Update $\mathbf{P}$}
 }
 }
 \caption{Proxy Matrix Optimization.}
 \label{alg:proxy}
\end{algorithm}

For realization in a deep learning framework, Algorithm \ref{alg:proxy} is converted into a fully connected layer which is then integrated into a neural network. The dimensionality reduction loss is back-propagated along with the loss corresponding to the network task to obtain the optimal projection matrix. PMO leverages the autograd routine in Pytorch to back-propagate through the SVD and hence removing the need for analytical gradient calculation.

\section{Experiments and Results}
\begin{figure}[H]
\centering
\includegraphics[width = 5cm]{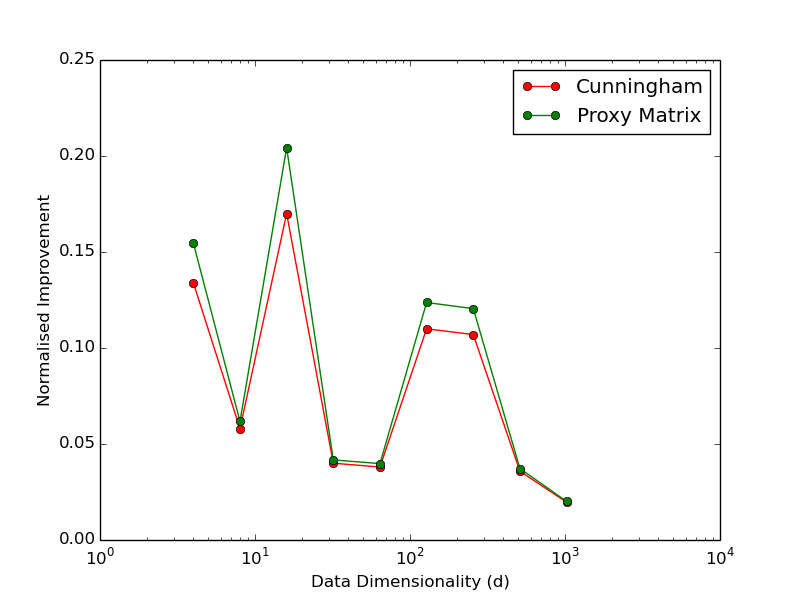}
\caption{Results from the $d$-sweep experiments using GILDA (with PMO) and two-step \cite{cunningham2015linear}.}
\label{fig:d-sweep}
\end{figure}

We perform two experiments to demonstrate the advantage of GILDA
and 
compare our results to the two-step method in \cite{cunningham2015linear} for the same toy dataset and initial conditions. For the first experiment, data of dimensionality, $d \in [4, 8, 16, 32, 64, 128, 256, 512, 1024]$ with $N = 2000$ points are generated. Each time the data is projected onto a space of dimension $r = 3$. The data is normally distributed with random covariance (exponentially distributed eccentricity with mean 2). The metric used to compare the methods is the normalised improvement over the eigenvector objective,  described as:
\begin{equation}
    -\frac{(f_X(\mathbf{R^{(orth)}})-f_X(\mathbf{R^{(eig)}}))}{|f_X(\mathbf{R^{(eig)}})|},
\end{equation}
where $f_X$ is the objective function described in Eq. \ref{eq:lda_obj}, $\mathbf{R^{(orth)}}$ is the projection matrix obtained using our PMO method and $\mathbf{R^{(eig)}}$ is the projection matrix obtained by a tradition eigenvector approach described in Section 2 by orthogonalizing the top $r$ eigenvectors of $\mathbf{\Sigma_W}^{-1} \mathbf{\Sigma_B}$. We also compare our approach to the two-step method provided by \cite{cunningham2015linear} using the same metric. The experiments were run 20 times and the results for the two methods are shown in Fig. \ref{fig:d-sweep}.
The results illustrate that for all the experiments, PMO outperforms \cite{cunningham2015linear} with the added benefit of not having to manually calculate the gradient of the objective function.

\begin{figure}[H]
\centering
\includegraphics[width = 5cm]{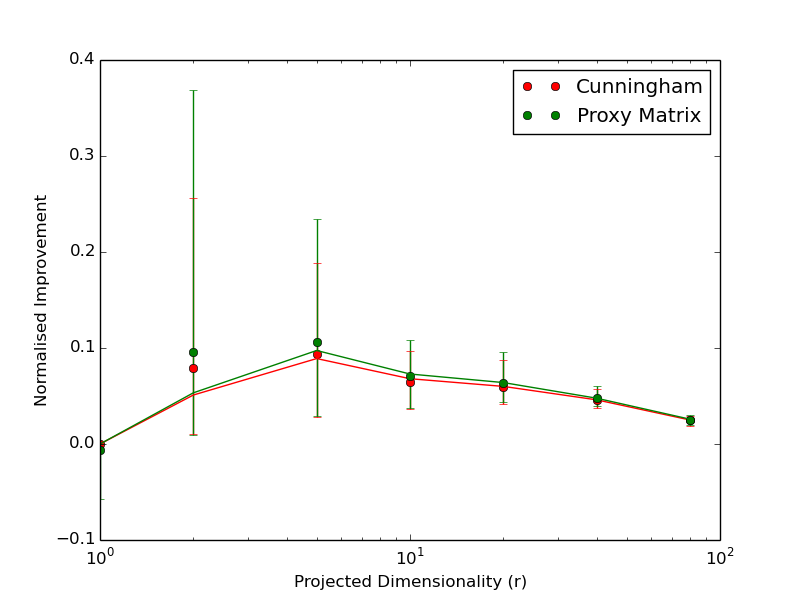}
\caption{Results from the $r$-sweep experiments 
using GILDA (with PMO) and two-step \cite{cunningham2015linear}.}
\label{fig:r-sweep}
\end{figure}

In the second experiment performed, the data dimensionality is fixed at $d = 100$ and the projected dimensionality takes on the values of $r \in [1, 2, 5, 10, 20, 40, 80]$ with $n = 1000$ data points in each of the $d$ classes. The within class data was generated according to a normal distribution with random covariance (uniformly distributed orientation and exponentially distributed eccentricity with mean 5), and each class mean vector was randomly chosen (normal with standard deviation 5/d). The results of the sweep are shown in Fig. \ref{fig:r-sweep}.
Apart form the case where $r = 1$, PMO does better than \cite{cunningham2015linear} and better than the eigenvector solution.

\section{Conclusion}
We introduced GILDA, a novel method for performing Linear Discriminant Analysis (LDA) using Proxy Matrix Optimization (PMO), that outperforms the popular two-step optimization method and heuristic solutions using eigenvectors. Furthermore, PMO is more functional as it leverages the inbuilt Pytorch SGD routine and automatic differentiation to find the gradients of the loss functions, and is suitable for realization within a deep learning framework. Future work would include integrating the LDA as a layer of a neural network to enable end-to-end training on real world datasets.

\section*{Acknowledgements}
This research was partly supported by the Air Force  Office  of  Scientific  Research  (AFOSR)  under  Dynamic  Data  Driven Applications  Systems  (DDDAS)  grant  FA9550-18-1-0121  and  the  National Science Foundation award number 1808582.


\bibliographystyle{IEEEtran}
\bibliography{neurips_2020}

\end{document}